\documentclass[journal]{IEEEtran}
\usepackage{multirow}
\usepackage{amssymb,amsmath}
\usepackage{rotating}
\usepackage[ruled,vlined]{algorithm2e}
\usepackage{algorithmic}
\usepackage{bm}
\usepackage{graphicx}
\usepackage{subfigure}
\usepackage{color}
\usepackage{amsthm}
\usepackage{amsbsy}

\usepackage{tabularx}
\usepackage{multirow}

\ifCLASSINFOpdf
\else
\fi
\begin{document}
%
\title{Age Group and Gender Estimation in the Wild with Deep RoR Architecture}
%
%
%

\author{Ke~Zhang,~\IEEEmembership{Member,~IEEE,}
        Ce~Gao,
        Liru~Guo,
        Miao~Sun,~\IEEEmembership{Student Member,~IEEE,}
        Xingfang~Yuan,~\IEEEmembership{Student Member,~IEEE,}
        Tony~X.~Han,~\IEEEmembership{Member,~IEEE,}
        Zhenbing~Zhao,~\IEEEmembership{Member,~IEEE，}
        and~Baogang~Li
\thanks{This work is supported by National Natural Science Foundation of China (Grants No. 61302163, No. 61302105, No. 61401154 and No. 61501185), Hebei Province Natural Science Foundation (Grants No. F2015502062, No. F2016502101 and No. F2016502062) and the Fundamental Research Funds for the Central Universities (Grants No. 2016MS99 and No. 2015ZD20).}
\thanks{K. Zhang is with the Department
of Electronic and Communication Engineering, North China Electric Power University, Baoding,
Hebei, 071000 China e-mail: zhangkeit@ncepu.edu.cn.}
\thanks{C. Gao is with the Department
  of Electronic and Communication Engineering, North China Electric Power University, Baoding,
  Hebei, 071000 China e-mail: 940770901@qq.com.}
\thanks{L. Guo is with the Department
  of Electronic and Communication Engineering, North China Electric Power University, Baoding,
  Hebei, 071000 China e-mail: glr9292@126.com.}
\thanks{M. Sun is with the Department
	of Electrical and Computer Engineering, University of Missouri, Columiba,
	MO, 65211 USA e-mail: msqz6@mail.missouri.edu.}
\thanks{X. Yuan is with the Department
	of Electrical and Computer Engineering, University of Missouri, Columiba,
	MO, 65211 USA e-mail: xyuan@mail.missouri.edu.}
\thanks{T. X. Han is with the Department
  of Electrical and Computer Engineering, University of Missouri, Columiba,
  MO, 65211 USA e-mail: HanTX@missouri.edu.}
\thanks{Z. Zhao is with the Department
	of Electronic and Communication Engineering, North China Electric Power University, Baoding,
	Hebei, 071000 China e-mail: zhaozhenbing@ncepu.edu.cn.}
\thanks{B. Li is with the Department
  of Electronic and Communication Engineering, North China Electric Power University, Baoding,
  Hebei, 071000 China e-mail: baogangli@ncepu.edu.cn.}
\thanks{Manuscript received 28 Aug. 2017; revised XX XXX. 2017.}}

%
%

\markboth{IEEE Transactions on \LaTeX\ Class Files,~Vol.~XX, No.~X, August~2017}%
{Shell \MakeLowercase{\textit{et al.}}: Bare Demo of IEEEtran.cls for IEEE Journals}
%



\maketitle

\begin{abstract}
Automatically predicting age group and gender from face images acquired in unconstrained conditions is an important and challenging task in many real-world applications. Nevertheless, the conventional methods with manually-designed features on in-the-wild benchmarks are unsatisfactory because of incompetency to tackle large variations in unconstrained images. This difficulty is alleviated to some degree through Convolutional Neural Networks (CNN) for its powerful feature representation. 
In this paper, we propose a new CNN based method for age group and gender estimation leveraging Residual Networks of Residual Networks (RoR), 
which exhibits better optimization ability for age group and gender classification than other CNN architectures.
Moreover, two modest mechanisms based on observation of the characteristics of age group are presented to further improve the performance of age estimation.
In order to further improve the performance and alleviate over-fitting problem, RoR model is pre-trained on ImageNet firstly, and then it is fune-tuned on the IMDB-WIKI-101 data set for further learning the features of face images, finally, it is used to fine-tune on Adience data set. 
Our experiments illustrate the effectiveness of RoR method for age and gender estimation in the wild, where it achieves better performance than other CNN methods. 
Finally, the RoR-152+IMDB-WIKI-101 with two mechanisms achieves new state-of-the-art results on Adience benchmark. 
\end{abstract}


%
\IEEEpeerreviewmaketitle

\section{Introduction}
%
%
%
%
\IEEEPARstart{A}{ge} and gender, two of the key facial attributes, play very foundational roles in social interactions, 
making age and gender estimation from a single face image an important task in intelligent applications, 
such as access control, human-computer interaction, law enforcement, marketing intelligence and visual surveillance, etc~\cite{Dex}. 

Over the last decade, most methods used manually-designed features and statistical models~\cite{Ma1,Ma2} to estimate age and gender~\cite{AgeGabor,AgeSfp,AgeBifSVMSVR,AgeLR,AgeSVR,AgePls,AgeCCA}, and they achieved respectable results on the benchmarks of {\em constrained} images, such as FG-NET~\cite{FG-NETAAM} and MORPH~\cite{Morph}. 
However, manually-designed features based methods behave unsatisfactorily on recent benchmarks of {\em unconstrained} images, namely ``in-the-wild'' benchmarks, including Public Figures~\cite{Pubfig}, Gallagher group photos~\cite{Gallagher}, Adience~\cite{AgeSVMdrop} and the apparent age data set LAP~\cite{LAP} for these features' ineptitude to approach large variations in appearance, noise, pose and lighting.
\par 
\begin{figure}
\centering
\includegraphics[width=1.0\linewidth]{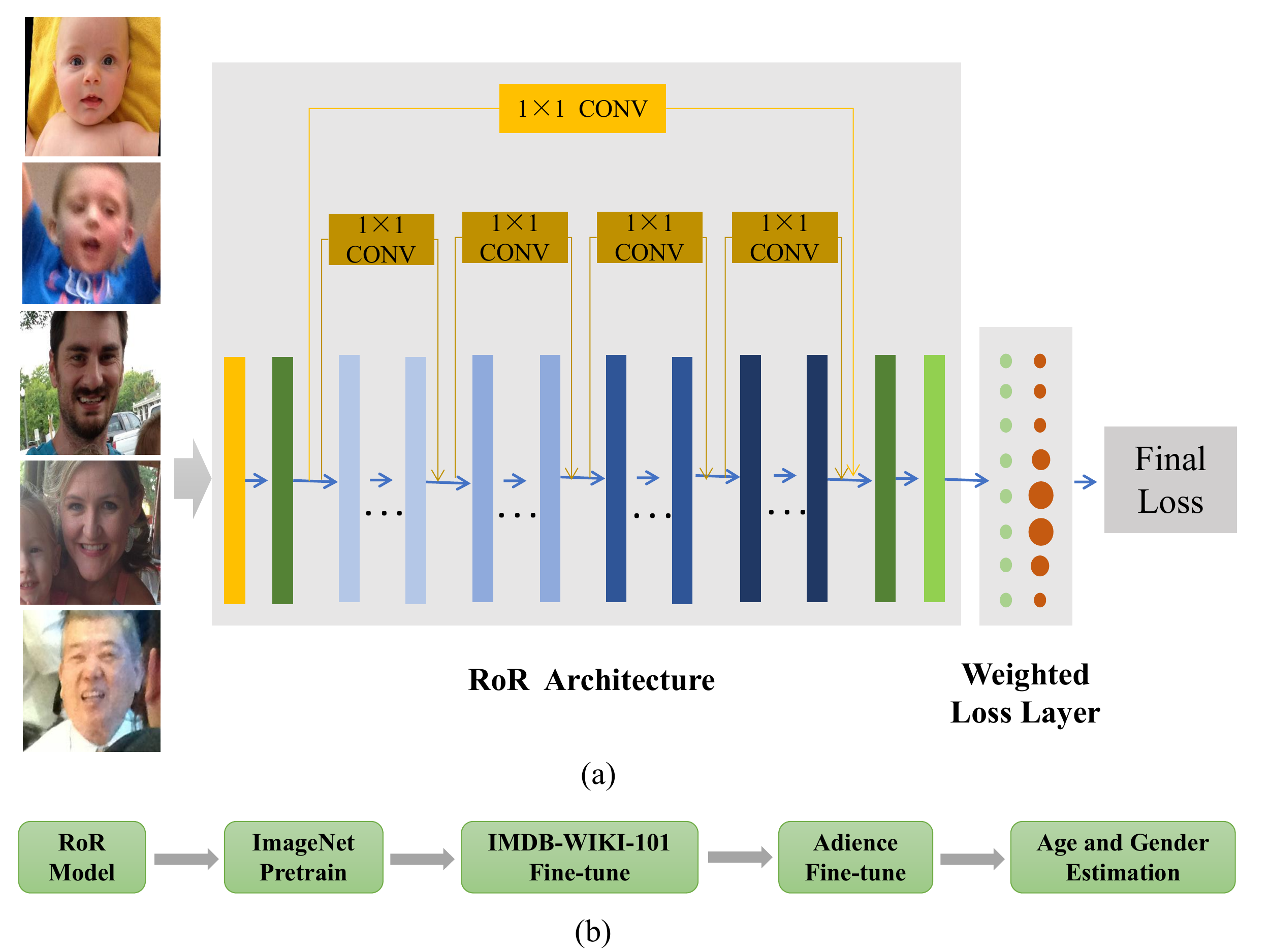}
\caption{Fig.1(a) is the overview of RoR architecture for age classification with weighted loss layer. The images from Adience data set represent some challenges of age and gender estimation from real-world, unconstrained images. RoR architecture is adopted for feature learning. In weighted loss layer, we use different loss weight instead of equal loss weight based on aging curve. The green circles stand for the original loss of every age group, and the red circles are denoted as different loss weight of every age group. Fig.1(b) is the pipeline of our framework. The RoR model is pre-trained on ImageNet firstly, and then it is fune-tuned on the IMDB-WIKI-101 data set for further learning the features of face images, finally, it is used to fine-tune on Adience data set for age and gender estimation.} 
\label{fig:pipe}
\end{figure}

Deep learning, especially deep Convolutional Neural Networks (CNN)~\cite{Alex,Miao,NIN,Overfeat,simonyan2014vgg,romero2014fitnets,lee2015dsn,springgenberg2014allcnn,szegedy2015googlenet,ResNet}, has proven itself to be a strong competitor to the more sophisticated and highly tuned methods~\cite{offtheshelf}. 
Although unconstrained photographic conditions bring about various challenges to age and gender prediction in the wild, we can still enjoy great improvements brought by CNNs~\cite{Agemulti,AgeWang,Agegenderbycnn,AgeSAAF,Dex}. 
The optimization ability of neural networks is critical to the performance of age and gender estimation, while existing CNNs designed for age and gender estimation only have several layers, which severely limit the development of age and gender estimation. Therefore, we construct a very deep CNN, Residual networks of Residual networks (RoR)~\cite{Ror}, for age group and gender estimation in the wild. 
To begin with, we construct RoR with different residual block types, and analyze the effects of drop-path, dropout, maximum epoch number, residual block type and depth in order to promote the learning capability of CNN.
In addition, analysis of the characteristics of age estimation suggests two modest mechanisms, pre-trained CNN by gender and weighted loss layer, to further increase the accuracy of age estimation, as shown in Fig.~\ref{fig:pipe}(a).
Moreover, in order to further improve the performance and alleviate over-fitting problem on small scale data set, we train RoR model on ImageNet firstly, and then fine-tune it on IMDB-WIKI-101 data set, thirdly, we use the model to further fine-tune on Adience data set. Fig.~\ref{fig:pipe}(b) shows the pipeline of our framework.
Finally, through massive experiments on Adience data set, our RoR model achieves the new state-of-the-art results on Adience data set. 
\par
The remainder of the paper is organized as follows. Section~\ref{sec2} briefly reviews related work for age and gender estimation methods and deep convolutional neural networks. The proposed RoR age and gender estimation method and the two mechanisms are described in Section~\ref{sec3}. Experimental results and analysis are presented in Section~\ref{sec4}, leading to conclusions in Section~\ref{sec5}.

\section{Related Work}\label{sec2}

\subsection{Age and gender estimation}
In the past twenty years, human age and gender estimation from face image has benefited tremendously from the evolutionary development in facial analysis. 
Early methods for age estimation were based on geometric features calculating ratios between different measurements of facial features~\cite{Kwon}. 
Geometry features can separate baby from adult easily but are unable to distinguish between adult and elderly people. 
Therefore, Active Appearance Model (AAM) based methods~\cite{FG-NETAAM} incorporated geometric and texture features to achieve desired result. 
However, these pixel-based methods are not suitable for in-the-wild images which have large variations in pose, illumination, expression, aging, cosmetics and occlusion.
After 2007, most existing methods used manually-designed features in this field, such as Gabor~\cite{AgeGabor}, LBP~\cite{AgeLbp}, SFP~\cite{AgeSfp}, and BIF~\cite{AgeBifSVMSVR}. Based on these manually-designed features, regression and classification methods are used to predict the age or gender of face images. 
SVM based methods~\cite{AgeBifSVMSVR,AgeSVMdrop} are used for age group and gender classification. For Regression, linear regression~\cite{AgeLR}, SVR~\cite{AgeSVR}, PLS~\cite{AgePls}, and CCA~\cite{AgeCCA} are the most popular methods for accurate age prediction. 
However, all of these methods were only proven effective on constrained benchmarks, and could not achieve respectable results on the benchmarks in the wild~\cite{shan2010learning,AgeSVMdrop}. 
\par 
Recent research on CNN showed that CNN model can learn a compact and discriminative feature representation when the size of training data is sufficiently large, so an increasing number of researchers start to use CNN for age and gender estimation. Yi et al.~\cite{Agemulti} first proposed a CNN based age and gender estimation method, Multi-Scale CNN. Wang et al.~\cite{AgeWang} extracted CNN features, and employed different regression and classification methods for age estimation on FG-NET and MORPH. Levi et al.~\cite{Agegenderbycnn} used CNN for age and gender classification on unconstrained Adience benchmark. Ekmekji~\cite{AgeEkmekji} proposed a chained gender-age classification model by training age classifiers on each gender separately. With the development of deeper CNNs, Liu et al.~\cite{Agenet} addressed the apparent age estimation problem by fusing two kinds of models, real-value based regression models and Gaussian label distribution based GoogLeNet on LAP data set. Antipov et al.~\cite{Aae} improved the previous year's results fusing general model and children model on LAP. Huo et al.~\cite{Dad} proposed a novel method called Deep Age Distribution Learning(DADL) to use the deep CNN model to predict the age distribution. Hou et al.~\cite{AgeSAAF} proposed a VGG-16-like model with Smooth Adaptive Activation Functions (SAAF) to predict age group on Adience benchmark. Then he used the exact squared Earth Mover’s Distance(EMD2)~\cite{EMD2} in loss function for CNN training and obtained better age estimation result. VGG-16 architecture and SVR~\cite{AgeRothe} were used for age estimation on top of the CNN features. Deep EXpectation (DEX) formulation~\cite{Dex} was proposed for age estimation based on VGG-16 architecture and a classification followed by a expected value formulation, and it got good results on FG-NET, MORPH, Adience and LAP data sets. Iqbal et al.~\cite{DAPP} proposed a local face description, Directional Age-Primitive Pattern(DAPP), which inherits discernible aging cue information and achieved higher accuracy on Adience data set. Recently, Hou et al. used the R-SAAFc2+IMDB-WIKI~\cite{SAAF+IMDB} method, and achieved the state-of-the-art results on Adience benchmark.

\subsection{Deep convolutional neural networks}
It is widely acknowledged that the performance of CNN based age and gender estimation relies heavily on the optimization ability of the CNN architecture, where deeper and deeper CNNs have been constructed.
From 5-conv+3-fc AlexNet~\cite{Alex} to the 16-conv+3-fc VGG networks~\cite{simonyan2014vgg} and 21-conv+1-fc GoogleNet~\cite{szegedy2015googlenet}, then to thousand-layer ResNets, both the accuracy and depth of CNNs were promptly increasing. With a dramatic rise in depth, residual networks (ResNets)~\cite{ResNet} achieved the state-of-the-art performance at ILSVRC 2015 classification, localization, detection, and COCO detection, segmentation tasks. Then in order to alleviate the vanishing gradient problem and further improve the performance of ResNets, Identity Mapping ResNets (Pre-ResNets)~\cite{he2016preresnets} simplified the residual networks training by BN-ReLU-conv order. Huang and Sun et al.~\cite{huang2016SD} proposed Stochastic Depth residual networks (SD), which randomly dropped a subset of layers and bypassed them with shortcut connections for every mini-batch to alleviate over-fitting and reduce vanishing gradient problem. In order to dig the optimization ability of residual networks family, Zhang et al.~\cite{Ror} proposed Residual Networks of Residual Networks architecture (RoR), which added shortcuts level by level based on residual networks, and achieved the state-of-the-art results on low-resolution image data sets such as CIFAR-10, CIFAR-100~\cite{krizhevsky2009cifar} and SVHN~\cite{netzer2011SVHN} at that time. Instead of sharply increasing the feature map dimension, PyramidNet~\cite{Pyramidal} gradually increases the feature map dimension at all units and gets superior generalization ability. DenseNet~\cite{Densenet} uses densely connected paths to concatenate the input features with the output features, and enables each micro-block to receive raw information from all previous micro-blocks. To enjoy the benefits from both path topologies of ResNets and DenseNet, Dual Path Network~\cite{Dpn} shares common features while maintaining the flexibility to explore new features through dual path architectures.

\section{Methodology}\label{sec3}
In this section, we describe the proposed RoR architecture with two modest mechanisms for age group and gender classification. Our methodology is essentially composed of four steps: Constructing RoR architecture for improving optimization ability of model, pre-training with gender and training with weighted loss layer for promoting the performance of age group classification, pre-training on ImageNet and further fine-tuning on IMDB-WIKI-101 data set for alleviating over-fitting problem and improving the performance of age group and gender classification. In the following, we describe the four main components in detail.
\subsection{Network architecture}
RoR~\cite{Ror} is based on a hypothesis: The residual mapping of residual mapping is easier to optimize than original residual mapping. To enhance the optimization ability of residual networks, RoR can optimize the residual mapping of residual mapping by adding shortcuts level by level based on residual networks. By experiments, Zhang et al.~\cite{Ror} argued that the optimization ability of Pre-RoR is better than RoR with the same number of layers, so we choose Pre-RoR in this paper except pre-training on ImageNet or IMDB-WIKI.
\par 
\begin{figure}
\centering
\includegraphics[width=1.0\linewidth]{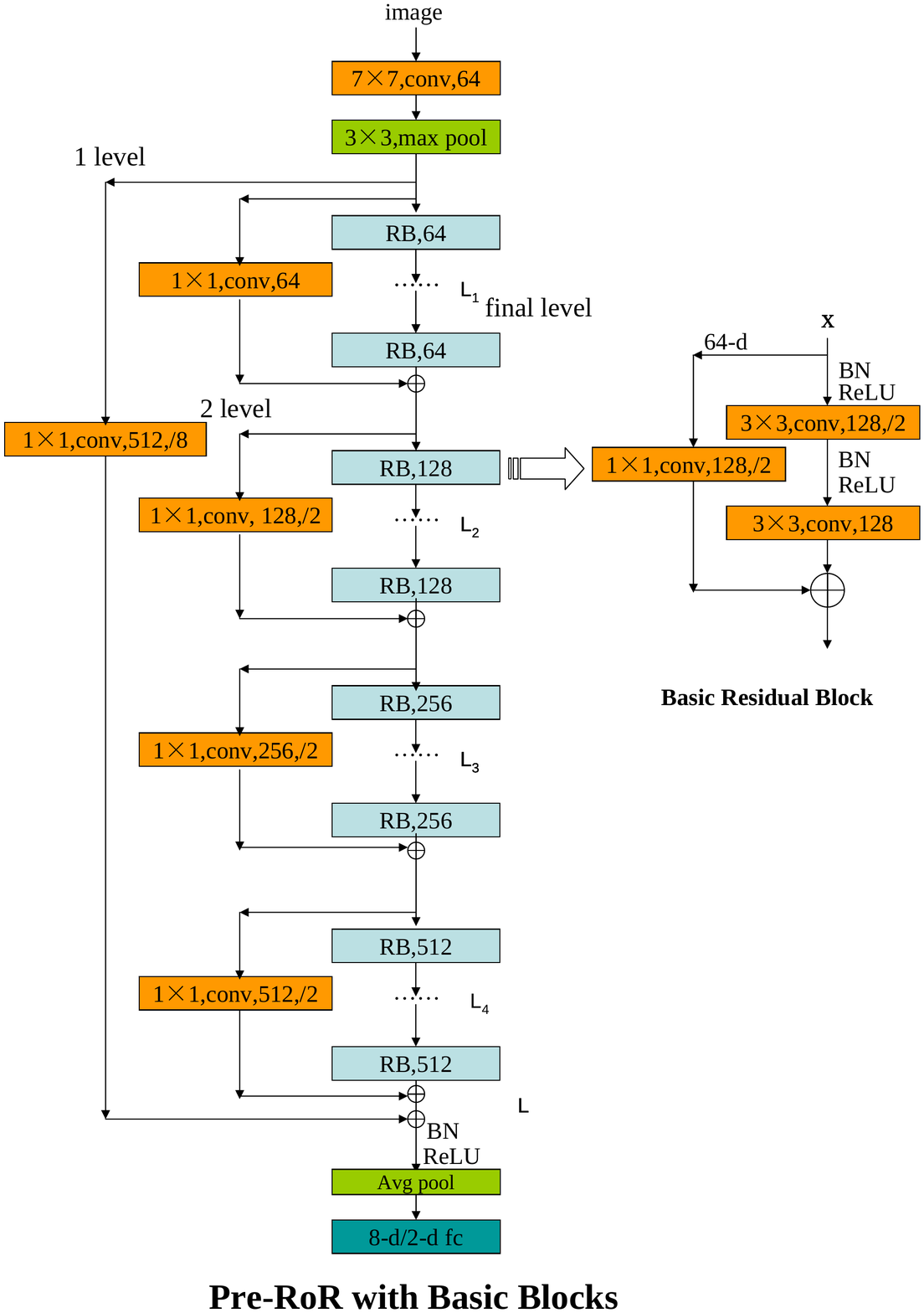}
\caption{Pre-RoR architecture with basic residual blocks. Pre-RoR has three levels, and it is constructed by adding shortcuts level by level based on basic Pre-ResNets. Leftmost shortcut is root-level shortcut, the rest four orange shortcuts are middle-level shortcuts, the blue shortcuts are final-level shortcuts. BN-ReLU-conv order in residual blocks is adopted. The fully-connected layer maps to the final soft-max layer for age or gender. Each basic residual block includes a stack of two convolutional layers.}
\label{fig:basic}
\end{figure}
In order to train the high-resolution Adience data set, we first construct RoR based on the basic Pre-ResNets for Adience, and denote this kind of RoR as Pre-RoR. Pre-ResNets~\cite{he2016preresnets} include two types of residual block designs: basic residual block and bottleneck residual block. 
Fig.~\ref{fig:basic} shows the Pre-RoR with basic block constructed based on original Pre-ResNets with $L$ basic blocks. The shortcuts in these $L$ original residual blocks are denoted as the final-level shortcuts.
 To start with, we add a shortcut above all basic blocks, and this shortcut can be called root shortcut or first-level shortcut. We use 64, 128, 256 and 512 filters sequentially in the convolutional layers, and each kind of filter has different number ($L_{1}, L_{2}, L_{3}, L_{4}$, respectively) of basic blocks which form four basic block groups. 
Furthermore, we add a shortcut above each basic block group, and these four shortcuts are called second-level shortcuts. Then we can continue adding shortcuts as the inner-level shortcuts. 
Lastly, the shortcuts in basic residual blocks are regarded as the final-level shortcuts. Let $m$ denote a shortcut level number. In this paper, we choose level number $m$=3 according to the analysis of Zhang et al.~\cite{Ror}, so the RoR has root-level, middle-level and final-level shortcuts, shown in Fig.~\ref{fig:basic}.
\par 
The junctions which are located at the end of each residual block group can be expressed by the following formulations.

\begin{equation} 
\begin{split}
x_{L_{1}+1}         =&g(x_{1})+h(x_{L_{1}})+F(x_{L_{1}}, W_{L_{1}}) \\
x_{L_{1}+L_{2}+1}   =&g(x_{L_{1}+1})+h(x_{L_{1}+L_{2}})+ \\ 
             & F(x_{L_{1}+L_{2}},W_{L_{1}+L_{2}}) \\
x_{L_{1}+L_{2}+L_{3}+1}  =&g(x_{L_{1}+L_{2}+1})+h(x_{L_{1}+L_{2}+L_{3}})+ \\
                        & F(x_{L_{1}+L_{2}+L_{3}}, W_{L_{1}+L_{2}+L_{3}}) \\
x_{L+1}   =&g(x_{1})+g(x_{L_{1}+L_{2}+L_{3}+1})+h(x_{L})+ \\
        & F(x_{L}, W_{L})
\end{split}
\end{equation} 
where $x_{l}$ and $x_{l+1}$ are input and output of the $l$-th block, and $F$ is a residual mapping function, $h(x_{l})=x_{l}$ and $g(x_{l})=x_{l}$ are both identity mapping functions. $g(x_{l})$ expresses the identity mapping of first-level and second-level shortcuts, and $h(x_{l})$ denotes the identity mapping of the final-level shortcuts. $g(x_{l})$ function is type B projection shortcut.
\par 

\begin{figure}
\centering
\includegraphics[width=1.0\linewidth]{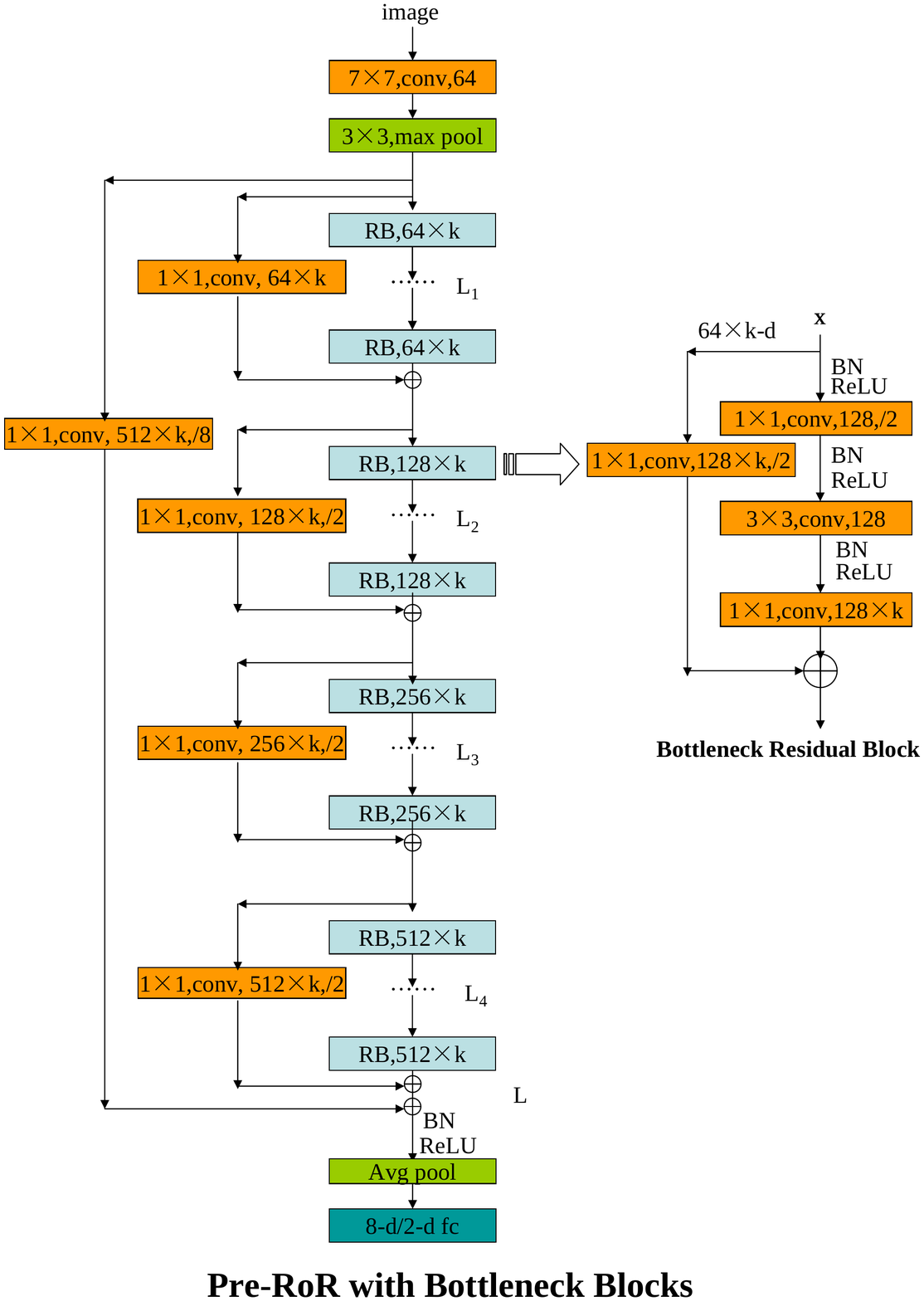}
\caption{Pre-RoR architecture with bottleneck residual blocks. If $k$=4, this is constructed based on original bottlencek Pre-ResNets architecture. The difference between this structure and Pre-RoR architecture with basic blocks is that its bottleneck block includes a stack of three convolutional layers.}
\label{fig:Pre-RoRnetworks}
\label{fig:bottleneck}
\end{figure}

For bottleneck block, He al et.~\cite{he2016preresnets} used a stack of three layers instead of two layers that first reduce the dimensions and then re-increase it. Both basic block and bottleneck block have similar time complexity, so we can get deeper networks easily through bottleneck. In this paper, we also construct a Pre-RoR based on bottleneck Pre-ResNets. The architecture details of Pre-RoR with bottleneck blocks are shown in Fig.~\ref{fig:bottleneck}. We use $k$ to control the output dimensions of the blocks. He et al.~\cite{he2016preresnets} chose $k$=4 led to the results that the input and output planes of these shortcuts are very different. 
Since the zero-padding (Type A) shortcut will bring more deviation and projection (Type B) shortcut will aggravate over-fitting, our RoR adopts $k$=4, $k$=2 and $k$=1 in this paper. 

\subsection{Pre-training with gender}

Like face recognition, age estimation can be easily affected by many intrinsic and extrinsic factors. Some of the most important factors include identity, gender and ethnicity, together with other factors like Pose, Illumination and Expression (PIE). We can alleviate the effects of these factors by using large data sets in the wild, but the existing data sets for age estimation are generally relatively small. To some extent, gender affects age judgments. On the one hand, the aging process of men slightly differs from women due to different longevity, hormones, skin thickness, etc. On the other hand, women are more likely to hide their real age by using makeup. So real-world age estimations for men and women are not exactly the same. 
Guo et al.~\cite{AgeCCA} and Ekmekji~\cite{AgeEkmekji} first manually separated the data set according to the gender labels, then trained an age estimator on each subset separately. 
Inspired by this, we train CNN by gender initially, then replace the gender prediction layer with age prediction layer, and fine-tune the whole CNN structure at last. 

\subsection{Training with weighted loss layer}

There are some diversities lying between general image classification and age estimation. Firstly, the different classes in general image classification are uncorrelated,
 but the age groups have a sequential relationship between labels. These interrelated age groups are more difficult to distinguish. Secondly, human aging processes show variations in different age ranges. For example, aging processes between mid-life adults and children are not equivalent. In this paper, we will analyze the law of human aging, and do age estimation under its guidance. For human, it is easier to distinguish who is the older one out of two people than to determine the persons' actual ages. Based on this characteristic and age-ordered groups, we define $y_{i}$, $i$={1,2...,$K$}, where $K$ is the number of age group labels. Then for a given age group $k\in K$, we separate the data set into two subsets $X_{k}^{+}$ and $X_{k}^{-}$ as follows:
\par
\begin{equation}
\begin{array}{cl}
X_{k}^{+}=\{(x_{i},+1)|y_{i}>k\} \\                     
X_{k}^{-}=\{(x_{i},+1)|y_{i}\le k\}
\end{array}
\label{E:2classes}
\end{equation}
\par
\begin{figure}

\centering
\includegraphics[width=1.0\linewidth]{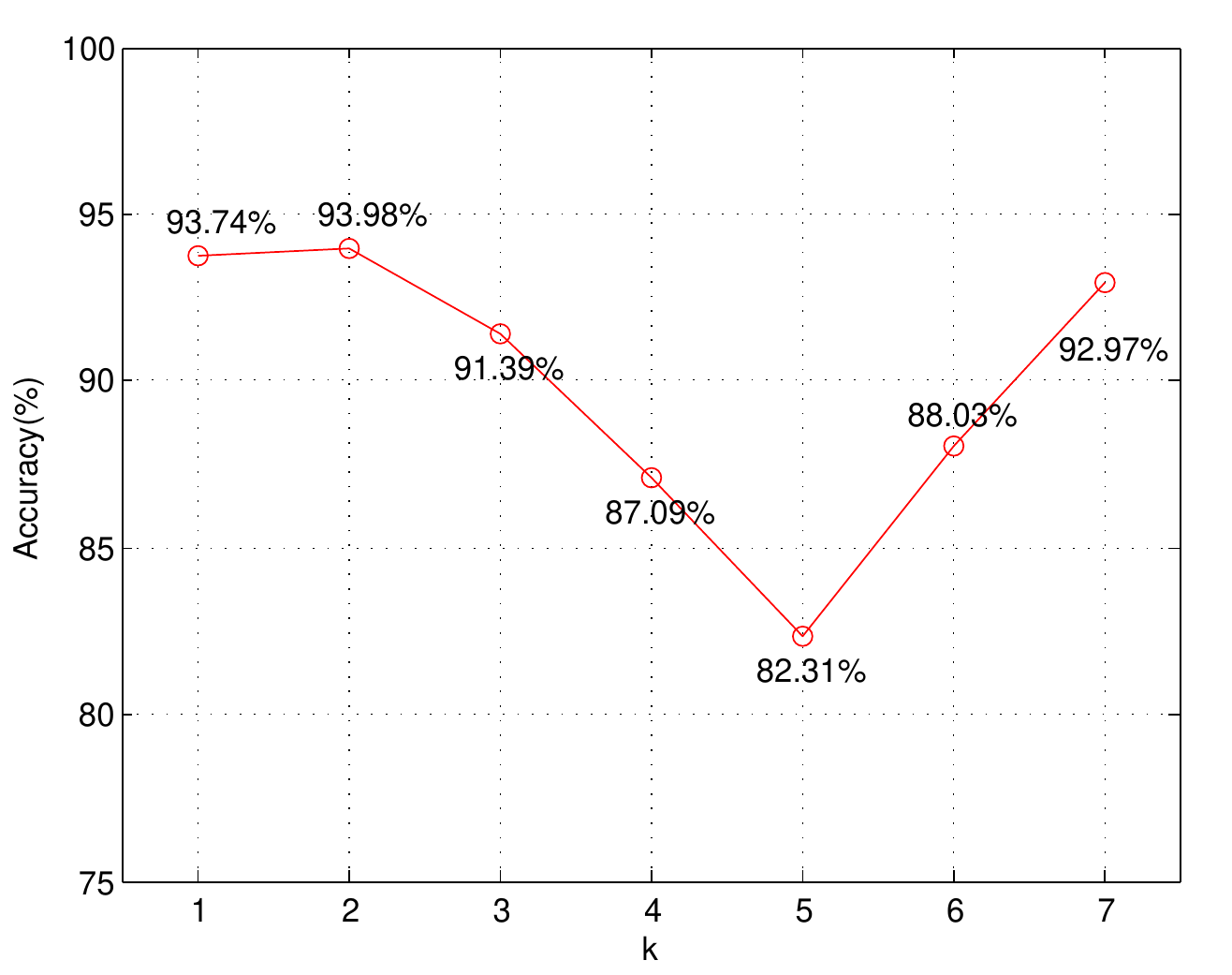}
\caption{The aging curve by binary classifiers. The curve expresses the aging rate. The lower the numerical value is, the more difficult it is to distinguish age group.}
\label{fig:agecurve}
\end{figure}

Next, we use the two subsets to learn a binary classifier that can be considered as a query: ``Is the face older than age group $k$?'' There are eight classes (0-2, 4-6, 8-13, 15-20, 25-32, 38-43, 48-53, 60-) in Adience data set, so we can choose $k$=1,2,...,7. By doing so, we get seven binary-class data sets, and the results of these binary classifiers can form a human aging curve which represents the human aging process. We execute some experiments on folder0 of Adience data set with 4c2f CNN described in~\cite{Agegenderbycnn} (just using two classes instead of eight classes), and the aging curve is described in Fig.~\ref{fig:agecurve} We discover that the 4th, 5th and 6th results are smaller than the others. As a conclusion, the aging process of smaller and greater age group is faster than intermediate age groups, so it is harder to distinguish intermediate age groups comparing to smaller and greater age groups.

\begin{table}[h]
\centering
\renewcommand{\arraystretch}{1.3}
\caption{Four different loss weight distributions.}
\begin{tabular}{|p{2.5cm}|p{4cm}|c|}
\hline
Name           &Loss Weight Distribution     \\ 
\hline
\hline
LW0                    &(1,1,1,1,1,1,1,1)     \\
\hline
LW1         &(1,1,1,0.9,0.8,0.8, 0.9,1)   \\
\hline
LW2         &(1,1,1,1.1,1.2,1.2,1.1,1)   \\
\hline
LW3           &(1,1,1,1.3,1.5,1.5,1.3,1)   \\
\hline
\end{tabular}

\label{tab:tab13}
\end{table}

Through above analysis, we realize the 4th, 5th, 6th and 7th groups are more difficult to estimate, so we apply higher loss weights to these age groups. 
Thus, we define four different settings of loss weight distributions for optimal results, as shown in Table~\ref{tab:tab13}.

\subsection{Pre-training on ImageNet}
Due to using small scale data sets for age and gender estimation, the over-fitting problem is easy to occur during training, so we use RoR network training ImageNet data set to obtain the basic feature expression model firstly. And then we use the pre-trained RoR model to fine-tune on the Adience data set, so as to alleviate the over-fitting problem brought by the direct training on Adience.
\par
The preceding data sets using RoR were all small scale image data sets, in this paper we first conduct experiments on large scale and high-resolution image data set, ImageNet. We evaluate our RoR method on the ImageNet 2012 classification data set~\cite{Russ2014imagenetchallenge}, which contains 1.28 million high-resolution training images and 50,000 validation images with 1000 object categories. During training of RoR, we notice that RoR is slower than ResNets. So instead of training RoR from scratch, we use the ResNets models from~\cite{gross2016facebookres} for pre-training. The weights from pre-trained ResNets models remain unchanged, but the new added weights are initialized as in~\cite{he2015prelu}. In addition, SD is not used here because SD makes RoR difficult to converge on ImageNet. Then we replace the 1000 classes prediction layer with age and gender prediction layer, and fine-tune the whole RoR structure on Adience. 

\subsection{Fine-tuning on IMDB-WIKI-101}
In order to make the RoR model further learn the feature expression of facial images and also reduce the over-fitting problem, we use large-scale face image data set IMDB-WIKI-101~\cite{Dex} to fine-tune the model after pre-training on ImageNet. 
\par
\begin{figure}
\centering
\includegraphics[width=1.0\linewidth]{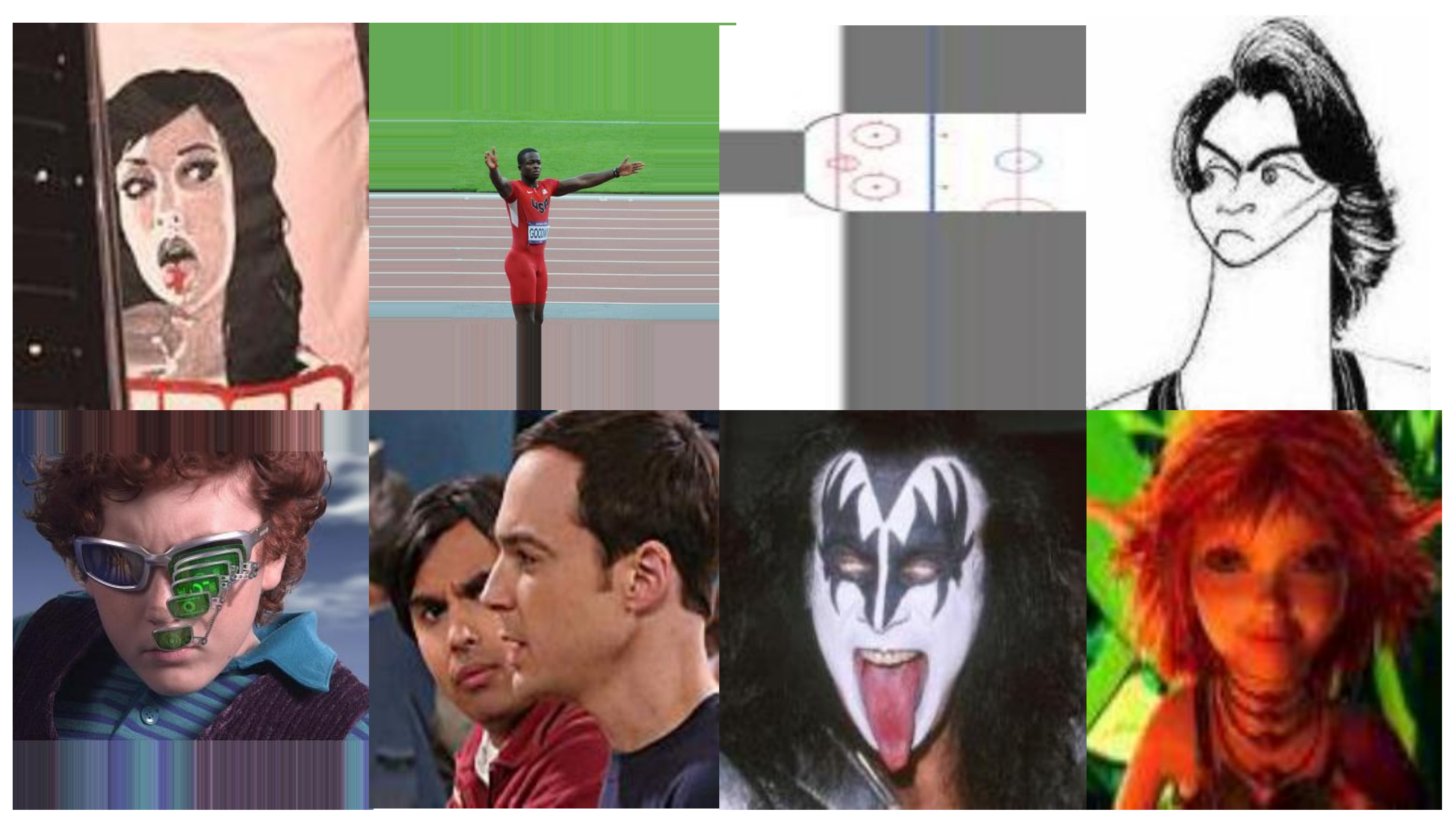}
\caption{The low-quality images in IMDB-WIKI.}
\label{fig:image2}
\end{figure}

IMDB-WIKI is the largest publicly available data set for age estimation of people in the wild, containing more than half million images with accurate age labels, whose age ranges from 0 to 100. For the IMDB-WIKI data set, the images were crawled from IMDb and Wikipedia, where IMDB contains 460723 images of 20,284 celebrities and Wikipedia contains 62328 images. As the images of IMDB-WIKI data set are obtained directly from the website, the IMDB-WIKI data set contains many low-quality images, such as human comic images, sketch images, severe facial mask, full body images, multi-person images, blank images, and so on. The example images are shown in Fig.~\ref{fig:image2}. Those bad images seriously affect the network learning effect. Therefore, in this paper, we spend a week manually removing the low quantity images by four people. In our removing process we mainly consider: a) the bad images, which are not standard face images from the IMDB-WIKI data set and b) the images with wrong age labels, especially the age images from 0 to 10 years old. The remaining IMDB-WIKI dataset remains 440607 images. The data set after cleaning is divided into 101 classes representing the age of each age, which we name IMDB-WIKI-101 data set. 
\par
Firstly, we replace the 1000 classes prediction layer on ImageNet with 101 classes prediction layer for age prediction, and fine-tune the RoR structure on IMDB-WIKI-101. When fine-tuning the RoR model, the IMDB-WIKI-101 data set is randomly divided into 90\% for training and 10\% for testing. Then we replace the 101 classes prediction layer with age and gender prediction layer, and fine-tune the whole RoR structure on Adience. 

\section{Experiments}\label{sec4}
In this section, extensive experiments are conducted to present the effectiveness of the proposed RoR architecture, two mechanisms, pre-training on ImageNet and further fine-tuning on IMDB-WIKI-101 data set. The experiments are conducted on unconstrained age group and gender data set, Adience~\cite{AgeSVMdrop}. Firstly, we introduce our experimental implementation. Secondly, we empirically demonstrate the effectiveness of two mechanisms for age group classification. Thirdly, we analyze different Pre-RoR models for age group and gender classification. Fourthly, we improve the performance of age and gender estimation by pre-training on ImageNet with RoR models. Furthermore, the RoR model are fine-tuned on IMDB-WIKI-101 data set for learning the feature expression of face images. Finally, the results of our best models are compared with several state-of-the-art approaches.
\subsection{Implementation}
For Adience data set, we do experiments by using 4c2f-CNN~\cite{Agegenderbycnn}, VGG~\cite{simonyan2014vgg}, Pre-ResNets~\cite{he2016preresnets}, our Pre-RoR architectures, respectively. \\
\textbf{4c2f-CNN}: The CNN structure described in~\cite{Agegenderbycnn} is denoted as baseline for the experiments with two mechanisms. Compared to the original 4c2f-CNN in~\cite{Agegenderbycnn}, our baseline adds preprocessing of data by subtracting the mean and dividing the standard deviation. 
\\\textbf{VGG}: We choose VGG-16 in~\cite{simonyan2014vgg} to construct age group and gender classifiers.
\\\textbf{Pre-ResNets}: We use Pre-ResNets-34, Pre-ResNets-50 and Pre-ResNets-101 in~\cite{he2016preresnets} as the basic architectures.
\\\textbf{Pre-RoR}: We use the basic block and bottleneck block Pre-ResNets in~\cite{he2016preresnets} 
to construct RoR architecture. The original Pre-ResNets contain four groups (64 filters, 128 filters, 256 filters and 512 filters) of residual blocks, the feature map sizes are 56, 28, 14 and 7, respectively. Pre-RoR with basic blocks includes Pre-RoR-34 (34 layers), Pre-RoR-58 (58 layers) and Pre-RoR-82 (82 layers). Pre-RoR with bottleneck blocks includes RoR-50 (50 layers) and RoR-101 (101 layers). Each residual block group in different Pre-RoR has different number of residual blocks, as shown in Table~\ref{tab:tab6}. Pre-RoR contains four middle-level residual blocks (every middle-level residual block contains some final-level residual blocks) and one root-level residual block (the root-level residual block contains four middle-level residual blocks). We adopt BN-ReLU-conv order, as shown in Fig.~\ref{fig:basic} and Fig.~\ref{fig:bottleneck}. 

\par
\begin{table}[h]
\caption{The number of residual blocks.}
\centering
\begin{tabular}{|c|p{1.5cm}|p{3cm}|}
\hline
Block Type       &Number of Layers   &Number of blocks in each Group   \\ 
\hline\hline
Basic Block      &34         &{3, 4, 6, 3}    \\
\hline
Basic Block  &58         &{5, 6, 12, 5}   \\
\hline
Basic Block  &82         &{7, 8, 14, 7}  \\
\hline
Bottleneck Block &50         &{3, 4, 6, 3}   \\
\hline
Bottleneck Block &101        &{3, 4, 23, 3}   \\
\hline
\end{tabular}

\label{tab:tab6}
\end{table}
Our implementations are based on Torch 7 with one Nvidia Geforce Titan X. We initialize the weights as in~\cite{ResNet}. We use SGD with a mini-batch size of 64 for these architectures except Pre-RoR with neckbottle block where we use mini-batch size 32. The total epoch number is 164. The learning rate starts from 0.1, and is divided by a factor of 10 after epoch 80 and 122. We use a weight decay of 1e-4, momentum of 0.9, and Nesterov momentum with 0 dampening~\cite{gross2016facebookres}. For stochastic depth drop-path method, we set $p _{l}$ with the linear decay rule of $p_{0}$= 1 and $p_{L}$=0.5~\cite{huang2016SD}.
\par 
The entire Adience collection includes 26,580 256$\times$256 
color facial images of 2,284 subjects, with eight classes of age groups and two classes of gender. Testing for both age and gender classification is performed using a standard five-fold, subject-exclusive cross-validation protocol, defined in~\cite{AgeSVMdrop}. We use the in-plane aligned version of the faces, originally used in~\cite{hassner2015effective}. For data augmentation, VGG, PreResNets and Pre-RoR use scale and aspect ratio augmentation~\cite{gross2016facebookres} instead of scale augmentation used in 4c2f-CNN.

\subsection{Effectiveness of two mechanisms}
\begin{figure}
\centering
\includegraphics[width=1.0\linewidth]{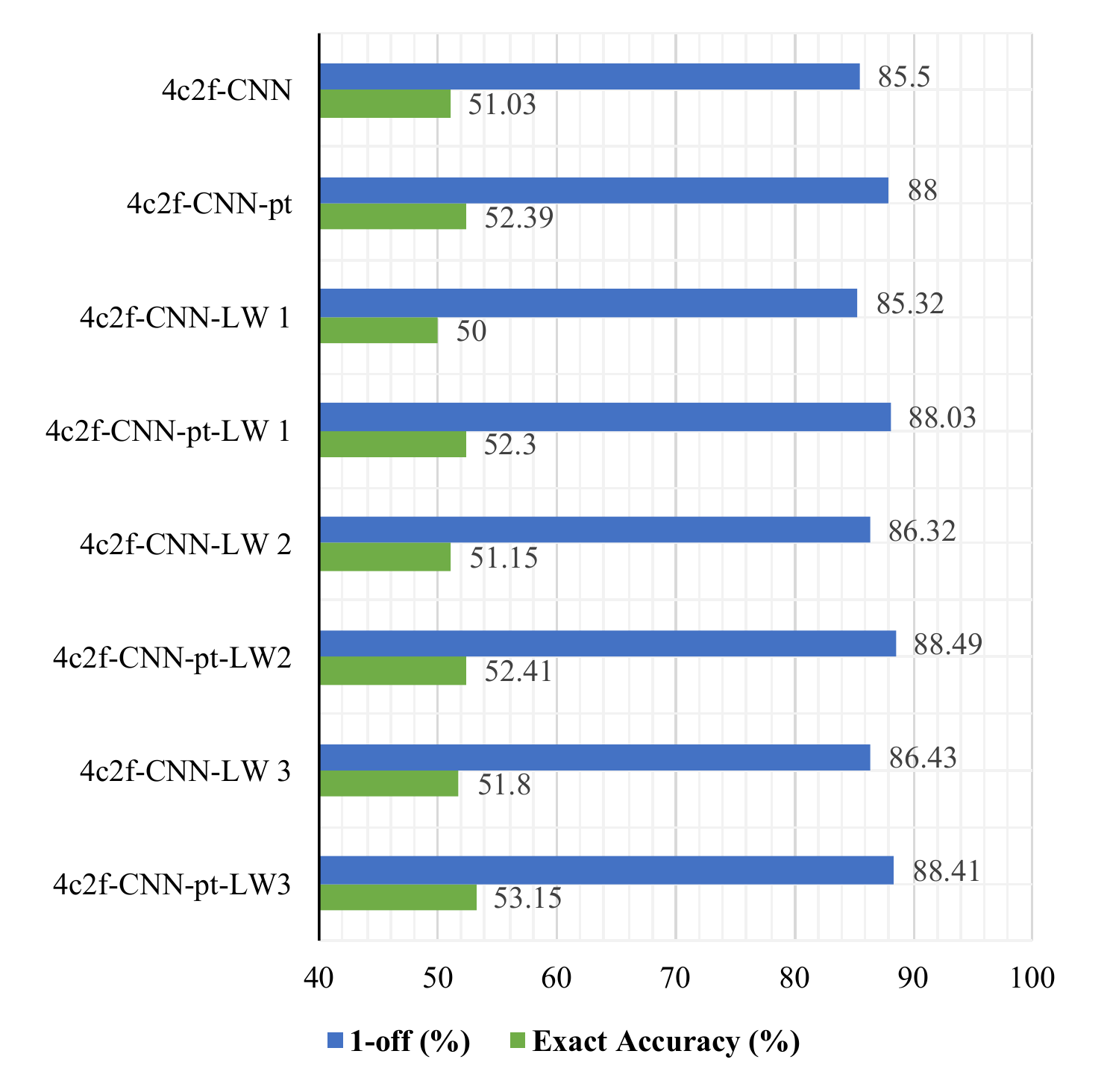}
\caption{Comparison of 4c2f-CNN and 4c2f-CNN with two mechanisms on folder0 of Adience.}
\label{fig:image4}
\end{figure}
In this section, we do age group classification experiments on folder0 of Adience data set with two mechanisms based on 4c2f-CNN architecture, and the results are described in Fig.~\ref{fig:image4}. Here, we report the \textbf{exact accuracy}(correct age group predicted) and \textbf{1-off accuracy} (correct or adjacent age group predicted) as~\cite{AgeSVMdrop}.


\par
Previously, we use 4c2f-CNN with each mechanism individually. In Fig.~\ref{fig:image4}, 4c2f-CNN pre-training by gender (4c2f-CNN-pt) achieves apparent progress compared to 4c2f-CNN without pre-training. And then, Fig.~\ref{fig:image4} also shows that 4c2f-CNN with loss weight distribution LW3 (4c2f-CNN-LW3) achieves best performance among all the loss weight distributions on folder0 of Adience data set, so we will choose LW3 as the loss weight distribution in the following experiments. Finally, we combine above the two mechanisms to predict age group and Fig.~\ref{fig:image4} shows that 4c2f-CNN combined of pre-training by gender and loss weight distribution LW3 together (4c2f-CNN-pt-LW3) achieves better performance than other models. These experiments demonstrate the effectiveness of pre-training method by gender and weighted loss layer for promoting performance of age group classification.

\subsection{Age group and gender classification by Pre-RoR}

In order to find the optimal model of Pre-RoR on Adience data set, we do a lot of comparative experiments with folder0 validation, and then we evaluate the effect of SD, dropout, shortcut type, block type, maximum epoch number and depth for age estimation results.
\par
\begin{figure}
\centering
\includegraphics[width=1.0\linewidth]{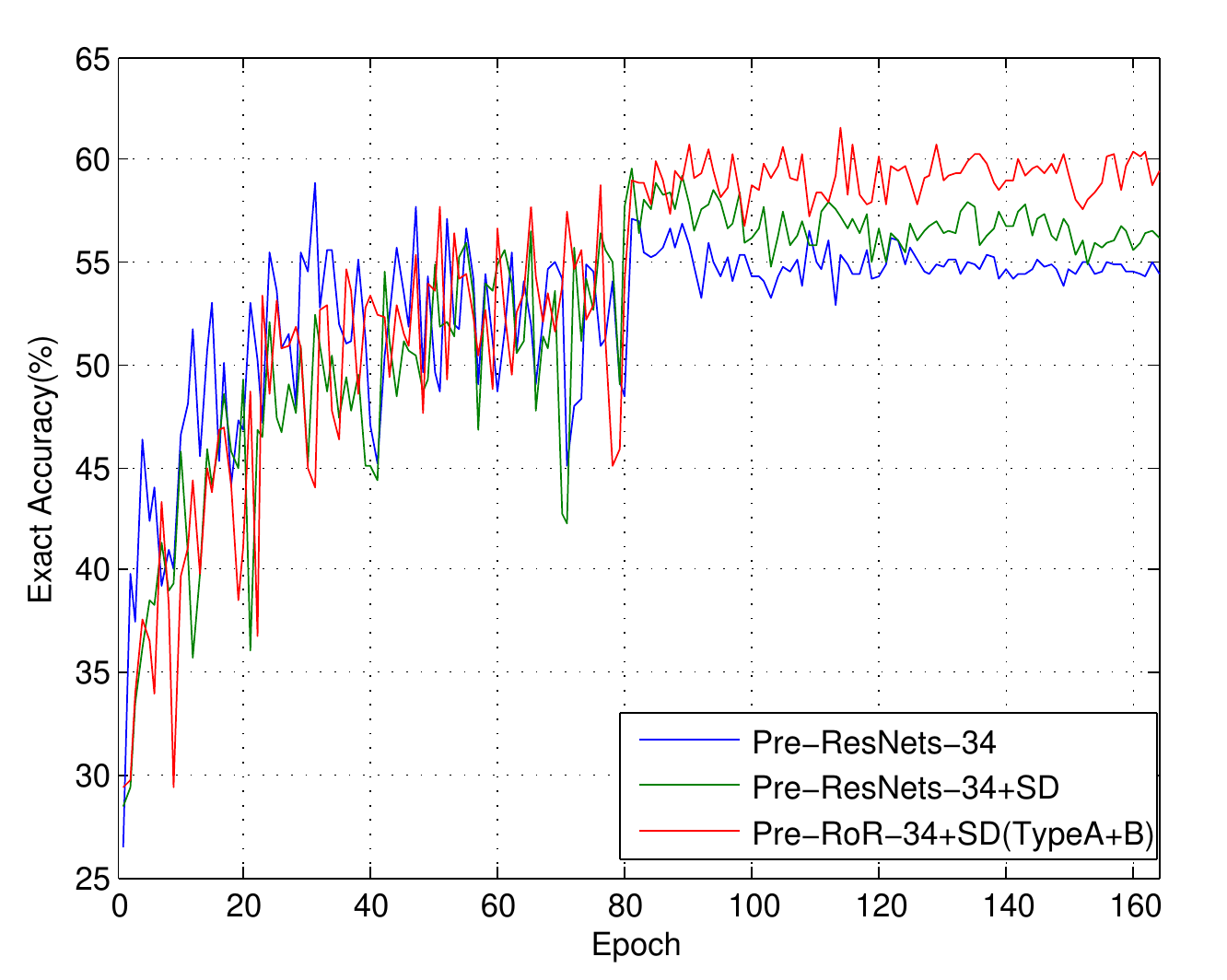}
\caption{Results on folder0 of Adience by Pre-ResNets-34, Pre-ResNets-34+SD and Pre-RoR-34+SD (Type A+B) during training, corresponding to results in Table~\ref{tab:tab7}. The blue curve of Pre-ResNets-34 shows that the over-fitting is very obvious. The green curve of Pre-RoR-34+SD) and the red curve of Pre-RoR-34+SD (Type A+B) shows the effectiveness of SD for reducing over-fitting. Pre-RoR-34+SD (Type A+B) displays stronger optimization ability of RoR.}
\label{fig:image5}
\end{figure}

 
\begin{table}[h]
\caption{Age and gender classification results on Adience benchmark with basic block architecture.}
\centering
\begin{tabular}{|p{2.9cm}|p{1.4cm}|p{1.1cm}|p{1.4cm}|}
\hline
Method  &Age Exact Accuracy(\%)  &Age 1-off(\%) &Gender Accuracy(\%)  \\ 
\hline\hline
Pre-ResNets-34 (Type B)       &58.81  &88.31  &90.23  \\
\hline
Pre-ResNets-34+SD (Type B)    &59.56  &90.43  &89.91  \\
\hline
Pre-RoR-34+SD (Type B)      &60.21  &91.14  &90.72  \\
\hline
Pre-RoR-34+SD+dropout (Type B)    &59.87  &88.68  &90.32  \\
\hline
Pre-RoR-34+SD  (Type A+B)       &61.56  &91.59  &90.78  \\
\hline
Pre-RoR-34+SD  (Type A+B) 300 epochs  &61.52  &91.56  &90.84  \\
\hline
Pre-RoR-58+SD (Type A+B)    &62.48  &92.31  &90.85  \\
\hline
Pre-RoR-82+SD (Type A+B)    &61.78  &92.15  &90.87  \\
\hline
\end{tabular}

\label{tab:tab7}
\end{table}
Firstly, basic blocks are used in experiments, and the results of different architectures are shown in Table~\ref{tab:tab7}. We do some experiments by Pre-ResNets-34 (34 convolutional layers) with and without SD. Because Adience data set only has about 26,580 high-resolution images, over-fitting is a critical problem. In Table~\ref{tab:tab7}, the performance of Pre-ResNets-34 with SD is better than that without SD, which means SD alleviates the effect of over-fitting. We then use Pre-RoR-34 +SD to estimate age and gender. Pre-RoR-34+SD outperforms Pre-ResNets-34+SD, because RoR can promote the learning capability of residual networks. To further reduce over-fitting, we try dropout between convolutional layers in residual blocks, but the result of Pre-RoR-34+SD+dropout shows that dropout method in RoR does not make a big difference. This is consistent with WRN~\cite{zagoruyko2016wrn}. Zhang et al.~\cite{Ror} noted that extra parameters would escalate over-fitting and the zero-padding (type A) would bring more deviation, so shortcut Type A should be used in the final-level and Type B should be used in other levels (called Type A+B). Table~\ref{tab:tab7} shows that the Pre-RoR-34+SD with Type A+B has better performance than Pre-RoR-34+SD which uses Type B in all levels. Fig.~\ref{fig:image5} shows that the test errors by Pre-ResNets-34, Pre-ResNets-34+SD and Pre-RoR-34+SD (Type A+B) at different training epochs with folder0 validation. Zhang et al.~\cite{Ror} proofed that maximum epoch number of 500 is necessary to optimize RoR on CIFAR-10 and CIFAR-100, but the results of Pre-RoR-34+SD with 300 epochs show that 164 for maximum epoch number is enough for Adience data set. Generally, ResNets~\cite{ResNet} and RoR~\cite{Ror} can improve performance by increasing depth. We estimate age and gender by Pre-RoR-58+SD and Pre-RoR-82+SD. The age estimation result of Pre-RoR-58+SD is better than Pre-RoR-34+SD, but Pre-RoR-82+SD is worse than Pre-RoR-58+SD, which is caused by degradation. Gender estimation gets better when adding more layers, since degradation is less critical for binary classification. 

Secondly, we use bottleneck blocks instead of basic blocks, and the results of different architectures are shown in Table~\ref{tab:tab8} and Table~\ref{tab:tab9}. We do some experiments by Pre-ResNets-50+SD (Type B, $k$=4) and Pre-RoR-50+SD (Type A+B, $k$=4). As can be observed, the performance of Pre-RoR-50+SD (Type A+B, $k$=4) is worse than Pre-ResNets-50+SD (Type B, $k$=4). When we use type A in final levels, the input and output planes of these shortcuts are very different, the zero-padding (type A) will bring more deviation. So we reduce the output dimensions by using $k$=2 and $k$=1.  The results of Pre-RoR-50+SD (Type A+B, $k$=2) and Pre-RoR-50+SD (Type A+B, $k$=1) show that deviation problem is largely alleviated by reducing dimensions. The performance of Pre-RoR-50+SD (Type A+B, $k$=2) is better than Pre-RoR-50+SD (Type A+B, $k$=1), because reducing dimensions also reduces parameters and the optimizing ability of networks. Pre-RoR-50+SD (Type A+B, $k$=2) achieves the balance of deviation and over-fitting problems, but it can not catch up Pre-RoR with basic blocks because of these two problems.

\begin{table}[h]
\caption{Age and gender classification results on Adience benchmark with 50-layer bottleneck block architecture.}
\centering
\begin{tabular}{|p{2.9cm}|p{1.5cm}|p{1.1cm}|p{1.3cm}|}
\hline
Method  &Age Exact Acc(\%)  &Age 1-off(\%) &Gender Acc(\%)   \\ 
\hline\hline
Pre-ResNets-50+SD (Type B) $k$=4      &60.05  &88.98  &89.82  \\
\hline
Pre-RoR-50+SD  (Type A+B)  $k$=4    &58.62  &90.10  &88.71  \\
\hline
Pre-RoR-50+SD  (Type A+B)  $k$=2    &61.68  &91.63  &88.92  \\
\hline
Pre-RoR-50+SD  (Type A+B)  $k$=1    &61.12  &91.14  &90.03  \\
\hline
\end{tabular}

\label{tab:tab8}
\end{table}

We do the same experiments by increasing the depth to 101 convolutional layers. We find the similar results shown in Table~\ref{tab:tab9} as the networks with 50 convolutional layers in Table~\ref{tab:tab8}. Pre-RoR-101+SD (Type A+B, $k$=2) achieves the best performance, and also outperforms Pre-RoR-50+SD (Type A+B, $k$=2).

\begin{table}[h]
\caption{Age and gender classification results on Adience benchmark with 101-layer bottleneck block architecture.}
\centering
\begin{tabular}{|p{3.1cm}|p{1.5cm}|p{1.1cm}|p{1.2cm}|}
\hline
Method  &Age Exact Acc(\%)  &Age 1-off(\%) &Gender Acc(\%)  \\ 
\hline\hline
Pre-ResNets-101+SD (Type B) $k$=4       &59.16  &89.61  &89.12  \\
\hline
Pre-RoR-101+SD  (Type A+B)  $k$=4   &60.46  &90.95  &88.37  \\
\hline
Pre-RoR-101+SD  (Type A+B)  $k$=2   &62.26  &91.54  &89.15  \\
\hline
Pre-RoR-101+SD  (Type A+B)  $k$=1   &60.49  &91.14  &89.41  \\
\hline
\end{tabular}

\label{tab:tab9}
\end{table}

In above experiments, we only use one folder to analyze different network architectures. Now we will demonstrate the generality of our method by using standard five-fold, subject-exclusive cross-validation protocol. In the following experiments, we only use Type A+B for Pre-RoR+SD. The age cross-validation results of Pre-RoR+SD (Type A+B) with different block types and depths are shown in Table~\ref{tab:tab10}, where we achieve the similar results with folder0 validation. The performance of Pre-RoR+SD with basic block is better than Pre-RoR+SD with bottleneck block. We analyze that this is because of deviation by zero-padding. 
Our Pre-ROR-58+SD achieves the best performance, which outperforms 4c2f-CNN by 18.8\% and 5.7\% on exact and 1-off accuracy of Adience data set.

\begin{table}[h]
\caption{The age cross-validation results of Pre-RoR with different block types and depths.}
\centering
\begin{tabular}{|p{3.2cm}|p{2.1cm}|c|}
\hline
Method &Exact Acc(\%)  &1-off(\%)  \\ 
\hline\hline
4c2f-CNN  &52.62$\pm$4.37   &88.61$\pm$2.27\\
\hline
VGG-16    &54.64$\pm$4.76 &54.64$\pm$4.76\\
\hline
Pre-ResNets-34  &60.15$\pm$3.99 &90.90$\pm$1.67\\
\hline
Pre-ResNets-34+SD &60.98$\pm$4.21 &91.87$\pm$1.73\\
\hline
Pre-RoR-50+SD $k$=2 &61.31$\pm$4.29 &93.45$\pm$1.34\\
\hline
Pre-RoR-50+SD $k$=1 &61.00$\pm$4.15 &93.19$\pm$1.67\\
\hline
Pre-RoR-101+SD $k$=2  &61.54$\pm$4.97 &93.37$\pm$1.72\\
\hline
Pre-RoR-101+SD $k$=1  &61.25$\pm$4.54 &93.52$\pm$1.59\\
\hline
Pre-RoR-34+SD           &62.35$\pm$4.69 &93.55$\pm$1.90\\
\hline
\textbf{Pre-RoR-58+SD}           &\textbf{62.50$\pm$4.33} &\textbf{93.63$\pm$1.90}\\
\hline
Pre-RoR-82+SD           &62.14$\pm$4.10 &93.68$\pm$1.22\\
\hline
\end{tabular}

\label{tab:tab10}
\end{table}

\subsection{Age group and gender classification by Pre-training on ImageNet}
Because we can not find the well-trained Pre-ResNets on the web, we construct RoR based on the well-trained ResNets from~\cite{gross2016facebookres} for ImageNet. The well-trained ResNets from~\cite{gross2016facebookres} use Type B in the residual blocks, so we use Type B in all levels of RoR. We use SGD with a mini-batch size of 128 (18 layers and 34 layers) or 64 (101 layers) or 48 (152 layers) for 10 epochs to fine-tune RoR. The learning rate starts from 0.001 and is divided by a factor of 10 after epoch 5. For data augmentation, we use scale and aspect ratio augmentation~\cite{gross2016facebookres}. Both Top-1 and Top-5 error rates with 10-crop testing are evaluated. From Table~\ref{tab:imagenet}, our implementation of residual networks achieves the best performance compared to ResNets methods for single model evaluation on validation data set. These experiments verify the effectiveness of RoR on ImageNet.
\begin{table}[!t]
\renewcommand{\arraystretch}{1.3}
\caption{Validation Error (\%, 10-crop testing) on ImageNet by ResNets and RoR with Different Depths}
\label{tab:imagenet}
\centering
\begin{tabular}{|l|c|c|}
\hline
Method                                       &Top-1 Error             &Top-5 Error  \\ \hline\hline
ResNets-18~\cite{gross2016facebookres}       &28.22                                  &9.42               \\\hline
\textbf{RoR-18}                            &\textbf{27.84}                                   &\textbf{9.22}                \\\hline
ResNets-34~\cite{ResNet}              &24.52                                  &7.46               \\\hline
ResNets-34~\cite{gross2016facebookres}       &24.76                                  &7.35               \\\hline
\textbf{RoR-34}                            &\textbf{24.47}                                   &\textbf{7.13}                \\\hline
ResNets-101~\cite{ResNet}             &21.75                                  &6.05               \\\hline
ResNets-101~\cite{gross2016facebookres}      &21.08                                  &5.35               \\\hline
\textbf{RoR-101}                           &\textbf{20.89}                                   &\textbf{5.24}               \\\hline
ResNets-152~\cite{ResNet}             &21.43                                  &5.71               \\\hline
ResNets-152~\cite{gross2016facebookres}      &20.69                                  &5.21               \\\hline
\textbf{RoR-152}                           &\textbf{20.55}                                   &\textbf{5.14}                \\\hline
\end{tabular}
\end{table}
\par
When we use pre-trained RoR model to fine-tune on Adience, we replace the 1000 classes prediction layer with age or gender prediction layer. We use SGD with a mini-batch size of 64 for 120 epochs to fine-tune on Adience. The learning rate starts from 0.01 and is divided by a factor of 10 after epoch 80. Based on the analysis of above section, we find deep Pre-RoR maybe outperform very deep Pre-RoR, so we use RoR-34 instead of deeper RoR as the basic pre-trained model. The results of different methods are shown in Table~\ref{tab:agegenderimagenet}. We do some experiments by ResNets-34 and RoR-34. The results of ResNets-34 and RoR-34 by Pre-training on ImageNet are better than the results of ResNets-34 and RoR-34, because pre-training on ImageNet can reduce over-fitting problem. When we add SD method in these experiments, the performance are promoted too. Especially, RoR-34+SD by Pre-training on ImageNet achieves very competitive performance, which outperforms Pre-RoR-34+SD. These experiments verify the effectiveness of pre-training on ImageNet for age group and gender classification. 
\begin{table}[!t]
\renewcommand{\arraystretch}{1.3}
\caption{Age group and gender classification results on Adience benchmark with RoR-34 by Pre-training on ImageNet}
\label{tab:agegenderimagenet}
\centering
\begin{tabular}{|p{1.8cm}|c|c|c|}
\hline
Method                                       &Age Exact Acc(\%)            &Age 1-off(\%)                 &Gender Acc(\%) \\ \hline\hline
ResNets-34                                   &59.39$\pm$4.45               &91.98$\pm$1.57                &90.12$\pm$1.48                            \\\hline
ResNets-34 by Pre-training on ImageNet        &61.15$\pm$4.53               &92.90$\pm$1.98                &91.18$\pm$1.53                            \\\hline
ResNets-34+SD by Pre-training on ImageNet     &61.47$\pm$5.17               &93.39$\pm$1.95                &91.98$\pm$1.49                            \\\hline
RoR-34                                   &60.29$\pm$4.25               &92.44$\pm$1.45                &91.07$\pm$1.64                            \\\hline
RoR-34 by Pre-training on ImageNet        &61.73$\pm$4.31               &92.97$\pm$1.55                &91.96$\pm$1.53                            \\\hline
\textbf{RoR-34+SD by Pre-training on ImageNet}     &\textbf{62.34$\pm$4.53}               &\textbf{93.64$\pm$1.47}                &\textbf{92.43$\pm$1.51}                            \\\hline

\end{tabular}
\end{table}

\subsection{Age group and gender classification by fine-tuning on IMDB-WIKI-101}
As the amount of training data strongly affects the accuracy of the trained models, there is a greater need for large datasets. Thus, we use IMDB-WIKI-101 to further fine-tune the RoR model. After pre-training on the ImageNet, we further fine-tune the RoR model on the IMDB-WIKI-101. The epoch is set to 120. The learning rate starts from 0.01 and is divided by a factor of 10 after epoch 60 and 90. When we use fine-tuned RoR model to fine-tune on Adience, we replace the 101 classes prediction layer with age or gender prediction layer. The epoch is set to 60. The learning rate is set to 0.0001.
\par
As shown in Table~\ref{tab:agegenderIMDB}, with the IMDB-WIKI-101 data set fine-tuning, both the performances of ResNets-34 and RoR-34 model have been significantly improved. This shows that having a large data set with face age images results in better performance. The performance of RoR-34 fine-tuning on the IMDB-WIKI-101 data set reaches the age exact accuracy of 66.74\%(1-off 97.38\%) compared to 60.29\% (1-off 92.44\%) when training directly on the Adience data set. That is competitive performance on Adience data set for age group and gender classification in the wild.
\par
When we only use ImageNet data set to pre-train the RoR-34 model, the age estimation results on Adience with stochastic depth algorithm are better than without stochastic depth algorithm. However, when we first use the ImageNet dataset to pre-train the RoR-34 network, and then use the IMDB-WIKI-101 data set to fine-tune the RoR-34 network, the age estimation results on the Adience with stochastic depth algorithm are worse than without stochastic depth algorithm. The reason is that the ImageNet dataset is an object image dataset, the network can learn the feature expression of general object, adding the stochastic depth algorithm to the original network is effective for the results. However, the IMDB-WIKI-101 is a large-scale face image data set. The RoR-34 network can fully learn the characteristics of face images from the IMDB-WIKI-101 data set, which reduces the problem of over-fitting. After adding stochastic depth algorithm, the original structure of the network will be changed, so the network needs to relearn the characteristics of facial image parameters, that is the reason why the results with SD are not better than the results without SD.

\begin{table}[!t]
\renewcommand{\arraystretch}{1.3}
\caption{Age group and gender classification results on Adience benchmark with RoR-34 by Fine-tuning on IMDB-WIKI-101}
\label{tab:agegenderIMDB}
\centering
\begin{tabular}{|p{1.8cm}|c|c|c|}
\hline
Method                                       &Age Exact Acc(\%)            &Age 1-off(\%)                 &Gender Acc(\%) \\ \hline\hline
ResNets-34+  IMDB-WIKI                      &66.63$\pm$3.04               &97.20$\pm$0.65                &93.17$\pm$1.57                \\\hline
RoR-34+  IMDB-WIKI  +SD                       &66.42$\pm$2.64               &97.35$\pm$0.65                &92.90$\pm$1.76                            \\\hline
\textbf{RoR-34+  IMDB-WIKI}                           &\textbf{66.74$\pm$2.69}               &\textbf{97.38$\pm$0.65}                &\textbf{93.24$\pm$1.77}                \\\hline

\end{tabular}
\end{table}

\subsection{Comparisons with state-of-the-art results of age group and gender classification on Adience}

To begin with, we use 4c2f-CNN, VGG-16, Pre-ResNets, our RoR+SD by Pre-training on ImageNet and Pre-RoR+SD architectures to estimate gender. In addition, we use IMDB-WIKI-101 dataset to fine-tune the ResNets-34 and RoR-34 for gender estimation. The gender cross-validation results by different methods are shown in Table~\ref{tab:tab11}.  RoR-34+SD achieves a competitive accuracy 92.43\% by only pretraining on ImageNet, and RoR-34+IMDB-WIKI achieves the best accuracy 93.24\%, which outperforms 4c2f-CNN~\cite{Agegenderbycnn} by 6.44\%. 

\begin{table}[h]
\caption{The gender cross-validation results by different methods.}
\centering
\begin{tabular}{|l|c|c|c|}
\hline
Method  &Exact Accuracy(\%)  \\ 
\hline\hline
SVM-dropout~\cite{AgeSVMdrop}     &79.3$\pm$0.0  \\
\hline
4c2f-CNN~\cite{Agegenderbycnn}      &86.8$\pm$1.4 \\
\hline
4c2f-CNN    &87.50$\pm$1.56 \\
\hline
VGG-16      &88.36$\pm$1.69 \\
\hline
Pre-ResNets-34    &92.04$\pm$1.51 \\
\hline
Pre-RoR-50+SD $k$=2 &90.45$\pm$1.39 \\
\hline
Pre-RoR-50+SD $k$=1 &90.66$\pm$1.41 \\
\hline
Pre-RoR-101+SD $k$=2  &91.09$\pm$1.44 \\
\hline
Pre-RoR-101+SD $k$=1  &91.31$\pm$1.54 \\
\hline
Pre-RoR-34+SD   &92.18$\pm$1.51 \\
\hline
Pre-RoR-58+SD   &92.29$\pm$1.49 \\
\hline
Pre-RoR-82+SD   &92.37$\pm$1.52 \\
\hline
RoR-34+SD by Pre-training on ImageNet  &92.43$\pm$1.51 \\
\hline
\textbf{ResNets-34+ IMDB-WIKI}  &\textbf{93.17$\pm$1.57} \\
\hline
\textbf{RoR-34+ IMDB-WIKI}  &\textbf{93.24$\pm$1.77} \\
\hline
\end{tabular}

\label{tab:tab11}
\end{table}

\begin{figure}
\centering
\includegraphics[width=1.0\linewidth]{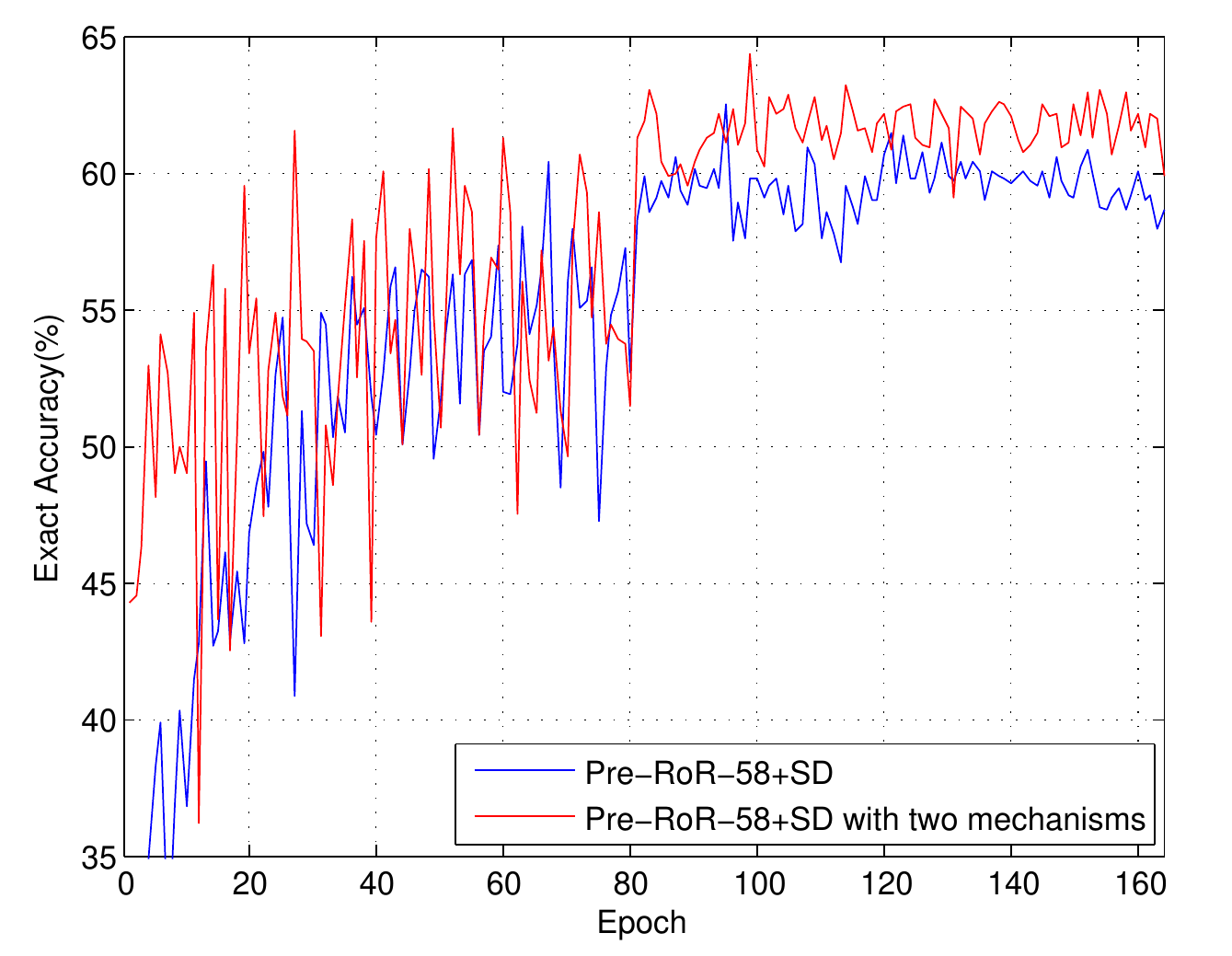}
\caption{Results on folder0 of Adience by Pre-RoR-58 and Pre-ROR-58+SD with two mechanisms during training. The red curve of Pre-ROR-58+SD with two mechanisms converges earlier and achieves higher accuracy than Pre-RoR-58.}
\label{fig:image6}
\end{figure}
Then, we use 4c2f-CNN, VGG-16, Pre-ResNets, our RoR-34+SD by Pre-training on ImageNet and Pre-RoR-58+SD (Type A+B) architectures with the two mechanisms to estimate age. Furthermore, we use IMDB-WIKI-101 dataset to fine-tune the ResNets-34 and RoR-34, and then with the two mechanisms for further age estimation on Adience.Table~\ref{tab:tab12} compares the state-of-the-art methods for age group classification on Adience data set. We find that the accuracy increases with the large-scale face image dataset fine-tuning the network, and two mechanisms will further improve each architecture, which demonstrates the versatility of two mechanisms in different models. Fig.~\ref{fig:image6} shows the test errors by Pre-ROR-58+SD and Pre-ROR-58+SD with two mechanisms at different training epochs with folder0 validation. In addition, we notice that the effect of RoR-34+IMDB-WIKI with two mechanisms is a little better than RoR-34+IMDB-WIKI without two mechanisms. We argue that this is because of well-trained model by IMDB-WIKI. 

\begin{table}[h]
\caption{The age cross-validation results by different methods.}
\centering
\begin{tabular}{|p{3.1cm}|c|c|}
\hline
Method &Exact Acc(\%)  &1-off(\%)  \\ 
\hline\hline
SVM-dropout~\cite{AgeSVMdrop}       &45.1$\pm$2.6 &79.5$\pm$1.4    \\
\hline
4c2f-CNN~\cite{Agegenderbycnn}      &50.7$\pm$5.1 &84.7$\pm$2.2   \\
\hline
Chained gender-age CNN~\cite{AgeEkmekji}    &54.5     &84.1           \\
\hline
R-SAAFc2~\cite{AgeSAAF}             &53.5     &87.9   \\
\hline
DEX w/o IMDB-WIKI pretrain~\cite{Dex}   &55.6$\pm$6.1 &89.7$\pm$1.8   \\
\hline
DEX w/ IMDB-WIKI pretrain~\cite{Dex}    &64.0$\pm$4.2 &96.60$\pm$0.90   \\
\hline
RES-EMD~\cite{EMD2}    &62.2   &94.3   \\
\hline
DAPP~\cite{DAPP}    &62.2 &--   \\
\hline
R-SAAFc2(IMDB-WIKI)~\cite{SAAF+IMDB}    &67.3   &97.0   \\
\hline
\hline
4c2f-CNN      &52.62$\pm$4.37 &88.61$\pm$2.27 \\
\hline
4c2f-CNN with two mechanisms    &53.96$\pm$3.80 &90.04$\pm$1.54 \\
\hline
VGG-16        &54.64$\pm$4.76 &89.93$\pm$1.87 \\
\hline
VGG-16 with two mechanisms    &56.11$\pm$5.05 &90.66$\pm$2.14 \\
\hline
Pre-ResNets-34      &60.15$\pm$3.99 &90.90$\pm$1.67 \\
\hline
Pre-ResNets-34 with two mechanisms  &61.89$\pm$4.16 &93.50$\pm$1.33 \\
\hline
Pre-RoR-58+SD       &62.50$\pm$4.33 &93.63$\pm$1.90 \\
\hline
Pre-RoR-58+SD with two mechanisms   &64.17$\pm$3.81 &95.77$\pm$1.24 \\
\hline
RoR-34+SD by Pre-training on ImageNet     &62.34$\pm$4.53               &93.64$\pm$1.47  \\
\hline
RoR-34+SD by Pre-training on ImageNet with two mechanisms  &63.76$\pm$4.18 &94.92$\pm$1.42 \\
\hline
\textbf{RoR-34+ IMDB-WIKI}   &\textbf{66.74$\pm$2.69} &\textbf{97.38$\pm$0.65} \\
\hline
\textbf{RoR-34+ IMDB-WIKI with two mechanisms}   &\textbf{66.91$\pm$2.51} &\textbf{97.49$\pm$0.76} \\
\hline
\textbf{RoR-152+ IMDB-WIKI with two mechanisms}   &\textbf{67.34$\pm$3.56} &\textbf{97.51$\pm$0.67} \\
\hline
\end{tabular}

\label{tab:tab12}
\end{table}

As shown in Table~\ref{tab:tab12}, without using ImageNet and IMDB-WIKI101 datasets, the accuracy of Pre-ROR-58+SD with two mechanisms is better than 64.0$\pm$4.2\% of DEX which pre-trained on ImageNet and IMDB-WIKI (523,051 face images)~\cite{Dex}. 
 Although DEX can achieve competitive results, it needs very large data set IMDB-WIKI for pre-training. Our method can learn age and gender representation from scratch without the IMDB-WIKI and achieve the best performance. Our VGG-16 with two mechanisms also outperforms DEX (also based on VGG-16) which only pre-trained on ImageNet but without IMDB-WIKI. These results demonstrate that our method can improve the optimization ability of networks and alleviate over-fitting on Adience data set. Moreover, by pre-training on ImageNet RoR-34+SD with two mechanisms also achieves 63.76$\pm$4.18\% of accuracy, which is very close to the accuracy in~\cite{Dex}, so we have reason to believe that better performance can be achieved by pre-training on more extra data sets. Particularly, our RoR-34+IMDB-WIKI with two mechanisms obtains a single-model accuracy of 66.91$\pm$2.51\% , and the 1-off accuracy of 97.49$\pm$0.76\% on Adience. But the single-model accuracy is slightly lower than the accuracy in~\cite{SAAF+IMDB}. Because compared with VGG used in~\cite{SAAF+IMDB} RoR-34 is small. So we use RoR-152+IMDB-WIKI to repeat the experiments, we get the new state-of-the-art performance (a single-model accuracy of 67.34$\pm$3.56\%) to our best knowledge now.

\section{Conclusion}\label{sec5}

This paper proposes a new Residual networks of Residual networks (RoR) architecture for high-resolution facial images age and gender classification in the wild. Two modest mechanisms, pre-training by gender and training with weighted loss layer, are used to improve the performance of age estimation. Pre-training on ImageNet is used to alleviate over-fitting. Further fine-tuning on IMDB-WIKI-101 is for the purpose of learning the features of face images. By RoR or Pre-RoR with two mechanisms, we obtain new state-of-the-art performance on Adience data set for age group and gender classification in the wild. Through empirical studies, this work not only significantly advances the age group and gender classification performance, but also explores the application of RoR on large scale and high-resolution image classifications in the future.


%

\appendices

\section*{Acknowledgment}
The authors would like to thank the editor and the anonymous reviewers for their careful reading and valuable remarks.

\ifCLASSOPTIONcaptionsoff
  \newpage
\fi

\end{document}